\documentclass[10pt,twocolumn,letterpaper]{article}

\usepackage[pagenumbers]{cvpr} 

\usepackage{graphicx}
\usepackage{amsmath}
\usepackage{amssymb}
\usepackage{booktabs}

\graphicspath{{graphics/}}
\usepackage{amsmath}
\usepackage{dsfont}
\usepackage{multirow}
\usepackage{wrapfig}
\usepackage{amsfonts}
\usepackage{booktabs}       
\usepackage{nicefrac}       
\usepackage{microtype}      
\usepackage{xcolor}         

\usepackage[pagebackref,breaklinks,colorlinks]{hyperref}

\usepackage[capitalize]{cleveref}
\crefname{section}{Sec.}{Secs.}
\Crefname{section}{Section}{Sections}
\Crefname{table}{Table}{Tables}
\crefname{table}{Tab.}{Tabs.}

\def\cvprPaperID{} 
\def\confName{CVPR}
\def\confYear{2023}

\begin{document}

\title{Towards Diverse Temporal Grounding under Single Positive Labels}

\author{Hao Zhou\textsuperscript{1}, Chongyang Zhang\textsuperscript{1,2}, Yanjun Chen\textsuperscript{1}, Chuanping Hu\textsuperscript{1,3}\\
\textsuperscript{1}School of Electronic Information and Electrical Engineering, Shanghai Jiao Tong University\\
\textsuperscript{2}MoE Key Lab of Artificial Intelligence, AI Institute, Shanghai Jiao Tong University, Shanghai, China\\
\textsuperscript{3}Zhengzhou University, Zhengzhou, China\\
{\tt\small zhouhaoah@outlook.com, sunny\_zhang@sjtu.edu.cn, erinchen92@163.com, cphu@vip.sina.com}
}
\maketitle

\begin{abstract}
Temporal grounding aims to retrieve moments of the described event within an untrimmed video by a language query. Typically, existing methods assume annotations are precise and unique, yet one query may describe multiple moments in many cases. Hence, simply taking it as a one-vs-one mapping task and striving to match single-label annotations will inevitably introduce false negatives during optimization. In this study, we reformulate this task as a one-vs-many optimization problem under the condition of single positive labels. The unlabeled moments are considered unobserved rather than negative, and we explore mining potential positive moments to assist in multiple moment retrieval. In this setting, we propose a novel Diverse Temporal Grounding framework, termed DTG-SPL, which mainly consists of a positive moment estimation (PME) module and a diverse moment regression (DMR) module. PME leverages semantic reconstruction information and an expected positive regularization to uncover potential positive moments in an online fashion. Under the supervision of these pseudo positives, DMR is able to localize diverse moments in parallel that meet different users. The entire framework allows for end-to-end optimization as well as fast inference. Extensive experiments on Charades-STA and ActivityNet Captions show that our method achieves superior performance in terms of both single-label and multi-label metrics.
\end{abstract}

\section{Introduction}
\label{sec:intro}
As the demand for video understanding increases, many related tasks have attracted a lot of attention, such as action recognition~\cite{2014Two,2016Temporal} and temporal action detection~\cite{Zhao_2017_ICCV,lin2017single}. Generally, these tasks rely on trimmed videos or predefined action categories. However, videos are untrimmed and accompanied by open-world language descriptions in most cases. Temporal grounding aims to localize corresponding temporal boundaries of the event in an untrimmed video by a language query, which is a crucial and challenging task in multi-modal understanding fields. Recently, this task has shown its potential in a wide range of applications, e.g. interest scene retrieval in movies.

\begin{figure*}[!t]
\begin{center}
   \includegraphics[width=\linewidth]{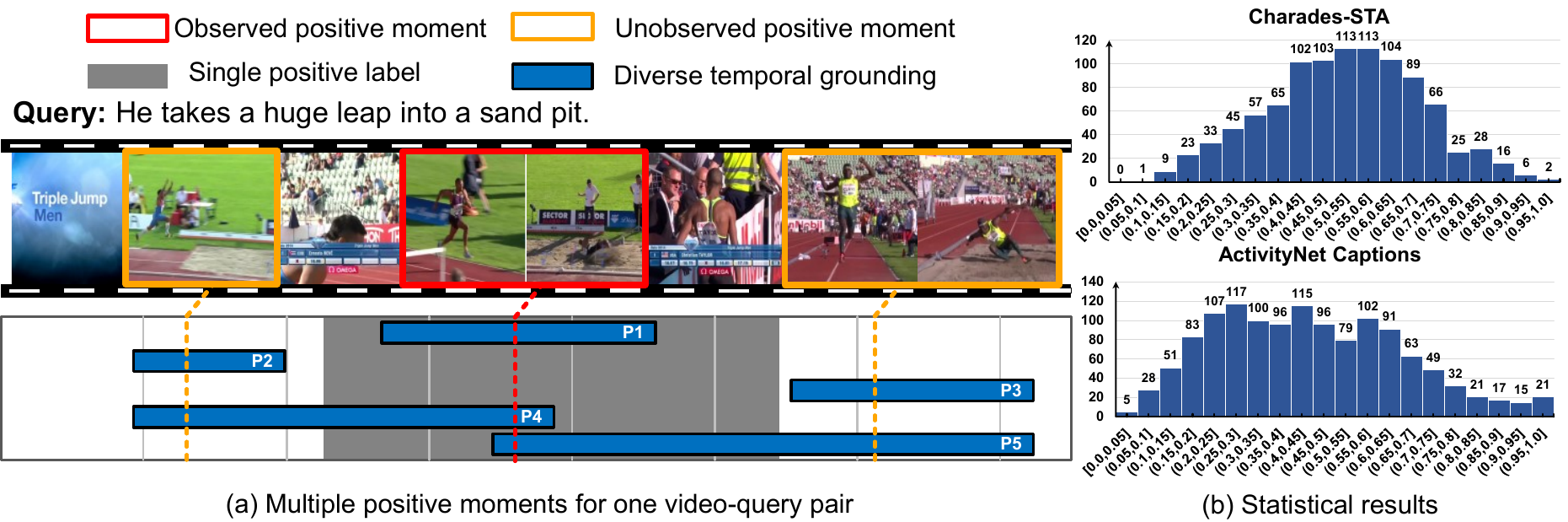}
\end{center}
   \caption{(a) The same query describes three similar moments (red and orange boxes) in the untrimmed video. Beyond matching the single positive label (the grey area), diverse temporal grounding aims to produce multiple predictions (blue bars) that meet the diversity of users and video contents. (b) Statistics of multiple annotations from partial test samples, re-collected by Otani~\etal\cite{otani2020uncovering}. X-axis denotes the average overlap (IoU) among five re-annotations for each video-query pair. Y-axis is the number of samples within various intervals. The results reveal that it is common for multiple moments to satisfy the same query.}
\label{fig:1}
\end{figure*}

Typically, temporal grounding is formulated as a one-vs-one mapping task, where each video-query pair is assumed to be associated with one positive label. As a result, most existing methods~\cite{zeng2020dense,yuan2020semantic} focus on better matching the single-label annotations through advanced multi-modal interaction mechanisms. Although great improvements have been achieved in recent years, we argue that this assumption may bring the risk of increasing error in the labels. In reality, a language query may describe multiple moments in a video, or annotators may provide multiple retrievals with the same query. Figure~\ref{fig:1}.b shows statistics of moment retrievals provided by multiple annotators. For a large proportion of samples, the average overlap among multiple annotations is less than 0.5. This phenomenon is widespread and inevitable because 1) {\bf repetition}: there are many repeated or similar moments in an untrimmed video, and 2) {\bf ambiguity}: temporal boundaries of the same event are subjective to users. When only one moment in a video-query pair is considered positive and the rest are negatives, many false negative labels are inevitably brought into the optimization procedure. Consequently, these false negatives will damage model generalization and cause a significantly performance drop~\cite{cole2021multi,zhou2022acknowledging}.

To alleviate the above optimization risk, a simple solution is to regard this task as multi-label optimization. However, these multi-label annotations are time-consuming and labor-intensive, so they are usually unavailable in realistic scenarios. Recently, many researchers~\cite{verelst2022spatial,li2022siod} have realized that data is imperfect in the real world and fully-observed labels are not accessible. We only have partial labels, \eg single positive labels, and the rest are unobserved labels in most cases. Inspired by this, we instead reformulate this task as a one-vs-many optimization problem under a condition of single positive labels. For each video-query pair, the annotated timestamp belongs to observed data as a single positive label, and the rest are treated as unobserved data rather than negative ones. At the same time, we strive for multiple outputs to cover unobserved positive moments. This formulation brings two appealing advantages. First, we can mitigate the optimization risks of false negatives without multi-label annotations. Second, the one-vs-many optimization allows for diverse moment localization. As observed in Figure~\ref{fig:1}.a, it is more applicable and user-friendly to retrieve diverse and plausible moments (P1-P5) instead of the most matched ones. To this end, the challenge becomes how to perform diverse temporal grounding from single positive labels.

In this paper, we present a novel Diverse Temporal Grounding framework under Single Positive Labels, called DTG-SPL, which mainly consists of a positive moment estimation (PME) module and a diverse moment regression (DMR) module. The PME uncovers unobserved positive moments in an online fashion, and these moments are then used as pseudo-positive labels for the transformer-based DMR module. Specifically, PME leverages a proposal-matching model to output matched scores for moment-query pairs in a video. In contrast to assuming that unobserved labels are negative, we apply a loss to only observed single positive labels. Meanwhile, an Expected Positive Regularization~\cite{cole2021multi} is applied to avoid the trivial "always predict positive" solution. Besides, PME introduces semantic information through a semantic-reconstruction model to further uncover potential positive moments. Based on an intuition that matched positive moments can better reconstruct the query, this model aims to generate both \textit{nouns} and \textit{verbs} that describe the segment-level content in a video. The whole framework can achieve end-to-end joint optimization. Finally, DTG-SPL is evaluated on two benchmarks, Charades-STA and ActivityNet Captions, and verified its superior performance for diverse temporal grounding. The contributions are summarized as follows:
\begin{itemize}
\item We reformulate temporal grounding as a one-vs-many optimization problem, by which we mitigate the optimization risk from false negatives and develop diverse temporal grounding.

\item We propose a novel diverse temporal grounding framework under single positive labels (DTG-SPL), which is able to mine unobserved positive moments for multi-label optimization in an online fashion.

\item We conduct extensive experiments on two public datasets, Charades-STA and ActivityNet Captions, and demonstrate that our method achieves state-of-the-art performance.
\end{itemize}

\section{Related Work}
\paragraph{Temporal Grounding}
Existing methods can be divided mainly into proposal-based and proposal-free frameworks.
Proposal-based methods view temporal grounding as a matching problem. Firstly, proposal-clip operations, \eg sliding windows, are applied to generate multi-scale proposals in an untrimmed video. These proposals are then fused separately with query features~\cite{gao2017tall,ning2021interaction}, or compared with them in the same latent space~\cite{hendricks2018localizing,anne2017localizing}, to obtain matched ones. To better interact multi-modal features, some methods~\cite{liu2018cross,yuan2020semantic} incorporate query semantics into language-temporal attention networks. Some methods~\cite{xu2019multilevel,lin2020moment} use query reconstruction as an auxiliary task to strengthen the cross-modal representations, whereas we use semantic reconstruction to estimate potential positives.
Proposal-free methods view temporal grounding as a regression or classification problem, where start-end timestamps are predicted directly~\cite{zeng2020dense,lu2019debug,ghosh2019excl}. Some methods~\cite{yuan2019find,mun2020local} build up attention-based location regression networks to enhance features of the labeled moment. Some methods~\cite{hahn2019tripping,he2019read} adopt reinforcement learning to intelligently skip around videos for efficient temporal grounding. Although great progress has been achieved, above methods follow the one-vs-one assumption and ignore the optimization risk of false negatives. Recently, some methods~\cite{rodriguez2020proposal,zhou2021embracing} leverage soft labeling technique or multiple-choice learning to alleviate this risk, yet still regard temporal grounding as one single-label optimization problem.

Recently, some researchers~\cite{lei2021detecting,chen2021end,bao2021dense,zhou2022thinking} introduce various transformer-based structures into the temporal grounding task. As one of the most related works, \cite{zhou2022thinking} utilizes a similar encoder-decoder architecture for multiple predictions. However, our proposed approach has significant advantages in two respects. Firstly, our approach does not require any additional training annotations, while~\cite{zhou2022thinking} relies on multiple annotations in the same video to generate soft multi-labels during optimization. Secondly, unlike \cite{zhou2022thinking}'s two-stage learning strategy, our method employs online positive moment estimation in a one-stage framework, resulting in a much lighter and faster training process.
\paragraph{Learning with Limited Labels}
In general, multi-label optimization requires full or multiple labels. However, most data is accompanied by limited labels and we cannot observe the fully-label information in realistic situations. Hence, many researchers investigate how to make the performance of models with limited labels as close as possible to that of fully-supervised models. One simple solution is to treat unobserved instances as negatives~\cite{bucak2011multi}, yet false negatives are brought at the same time. Recently, pseudo-label generation~\cite{tarvainen2017mean,kuo2020featmatch} has become a popular approach, where pre-trained networks are applied to predict unobserved labels.

Some methods~\cite{cole2021multi,verelst2022spatial,li2022siod} attempt to explore a more realistic setting, called single positive labels. Obviously, it is more challenging to train a multi-output model under single positive labels. Some researchers~\cite{cole2021multi} considers the expected number of positive labels per image as domain knowledge, and conducts online label estimation for the multi-label classification task. Some researchers~\cite{verelst2022spatial} presents a consistency loss to ensure that predictions are consistent over training epochs. Some researchers~\cite{li2022siod} explores a new single instance object detection task, where pseudo-label generation and contrastive learning are adopted to mine unobserved instances. Inspired by above works, we argue temporal grounding is closer to a one-vs-many optimization problem under single positive labels. 

\section{Proposed Approach}
\subsection{The Reformulation of Temporal Grounding}
For an untrimmed video $\mathrm{\bf V}$ and a language query $\mathrm{\bf Q}$ from the data space $\mathcal{D}$, temporal grounding aims to find a function $f \in \mathcal{F}$ that can predict the precise start-end boundary $\mathrm{\bf b}_{se}$. Specifically, the video is represented as $\mathrm{\bf V} = \{\mathrm{\bf x}^v_i\}_{i=1}^{T_c}$, where $\mathrm{\bf x}^v_i$ is the i-th video clip and $T_c$ is the total number of video clips. The query is represented as $\mathrm{\bf Q} = \{\mathrm{\bf x}^l_i\}_{i=1}^{T_l}$, where $\mathrm{\bf x}^l_i$ denotes the i-th word and $T_l$ is the total number of language words. The target can be represented as $\mathrm{\bf y} = \{1,0\}^{T_m}$, where $\mathrm{ y}_i = 1/0$ denotes the i-th moment is matched/unmatched and $T_m$ is the total number of moments in a video. Typically, $T_m$ is equal to the number of multi-scale proposals for the proposal-based frameworks. For the proposal-free frameworks, it is infinite depending on different start-end timestamps across the entire video. To this end, the goal is to find the function $f$ that minimizes the \textit{expected risk},
\begin{equation}
R(f,\mathrm{\bf y})  = \mathbb{E}_{(\mathrm{\bf Q},\mathrm{\bf V}) \sim \mathcal{D}}\mathcal{ L}(f(\mathrm{\bf Q},\mathrm{\bf V}),\mathrm{\bf y}),
\label{equ:1}
\end{equation}
where $\mathcal{ L}$ is a loss function that penalizes the differences between predictions and targets.

In practice, the data of temporal grounding task is imperfect and we cannot access to the fully-observed targets. For mainstream datasets, \eg Charades-STA and ActivityNet Captions, only single positive labels $\mathrm{\bf z}$ are available for each video-query pair, which are characterized by
\begin{equation}
\mathrm{\bf z} = \{1,\emptyset\}^{T_m}, \quad \sum\nolimits_{i=1}^{T_m} \mathds{1}[\mathrm{\bf z}_i = 1] = 1,
\label{equ:2}
\end{equation}
where $\mathrm{\bf z}_i = \emptyset$ denotes the label of i-th moment is unobserved, and $\mathds{1}[ \cdot ]$ is a binary indicator. However, prior works assume unobserved labels are negative and typically minimize the following \textit{risk},
\begin{equation}
\hat{R}(f,\mathrm{\bf z})  \!=\! \mathbb{E}_{(\mathrm{\bf Q},\mathrm{\bf V}) \sim \mathcal{D}}\mathcal{ L}(f(\mathrm{\bf Q},\mathrm{\bf V}),\mathrm{\bf z}), \forall \mathrm{\bf z}_i \!=\! \emptyset \Rightarrow \mathrm{\bf z}_i \!=\! 0.
\label{equ:3}
\end{equation}

Based on above assumption, multiple positive moments would be mistakenly treated as negative ones, thus leading to one optimization risk. To the best of our knowledge, we are the first to develop this task as diverse temporal grounding under single positive labels. Unlabeled moments are considered to be unobserved ones in this setting. Beyond localizing the observed single positive one, we attempt to retrieve multiple positive moments to meet the diversity of users and video contents. Formally, the new objective is to find a function that bridges the performance gap between partial-supervised and fully-supervised models,
\begin{equation}
\mathop{\mathrm{argmin}}_{\textit{f}} R(f,\mathrm{\bf z}) - R(f,\mathrm{\bf y}) .
\label{equ:4}
\end{equation}
\subsection{Method Overview}
\begin{figure*}[!t]
\begin{center}
   \includegraphics[width=\linewidth]{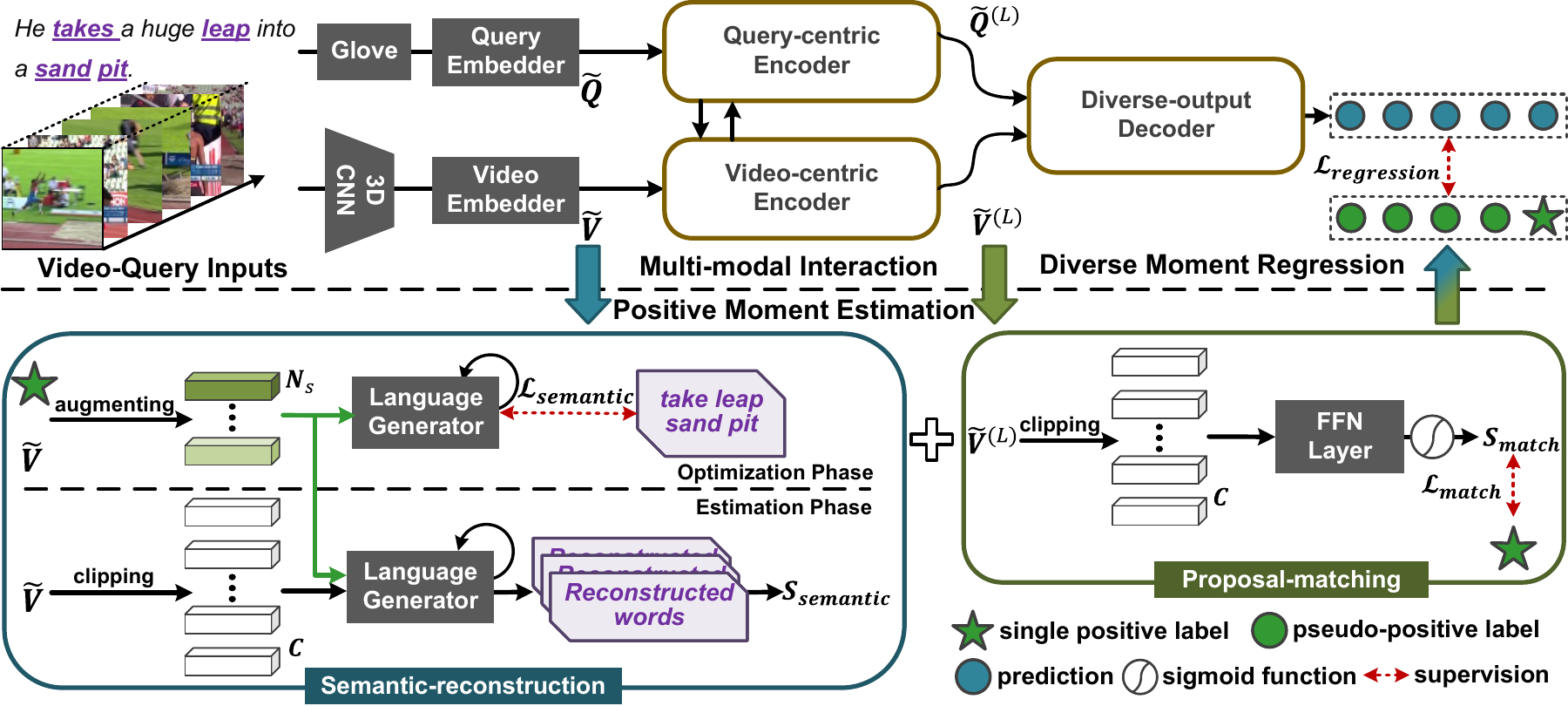}
\end{center}
   \caption{An overview of our proposed Diverse Temporal Grounding framework under Single Positive Labels (DTG-SPL). Dependent on collaboration of semantic-reconstruction and proposal-matching models, an online positive moment estimation module can uncover unobserved positive moments, which are then taken as pseudo-positive labels during optimization. }
\label{fig:2}
\end{figure*}
In order to bridge the performance gap in Equation~\ref{equ:4}, one of effective solutions is to find another function $g: (\mathcal{D},\mathrm{\bf z}) \rightarrow \mathrm{\bf y}$ that estimates labels of unobserved moments by single positive moments. Following this motivation, we present an online estimate network $g$ to mine potential positive moments for each training sample. These moments are labelled as pseudo-positives $\mathrm{\bf \Hat{y}}$, together with single positive labels $\mathrm{\bf z}$, to optimize one diverse temporal grounding network \textit{f}.

Figure~\ref{fig:2} illustrates our proposed Diverse Temporal Grounding framework under Single Positive Labels (DTG-SPL), including multi-modal interaction (Section \ref{MFE}), positive moment estimation module (dubbed PME, Section \ref{PME}), and diverse moment regression module (dubbed DMR, Section \ref{DMR}). Specifically, we first extract the sequential video and language features and adopt dual transformer-based encoders to allow for multi-modal interaction. PME is then applied to uncover multiple positive moments for each training sample under a single positive label. Finally, DMR is adopted to localize multiple moments. During the training process, DTG-SPL is optimized in an end-to-end manner. After each epoch, outputs of PME are used as pseudo-positive labels to guide the optimization of DMR in the next epoch. The overall loss is given by
\begin{equation}
\mathcal{ L}= \mathcal{ L}_{PME} + \mathcal{ L}_{DMR}.
\label{equ:5}
\end{equation}

Note that PME is not working during the inference process. To better validate the effectiveness of our proposed method, we further incorporate multi-label metrics to measure whether models can retrieve multiple positive moments successfully.
\subsection{Video-Language Feature Interaction}
\label{MFE}
\paragraph{Feature extraction}
We first extract sequential features from raw videos and language queries. Specifically, a pre-trained 3D CNN model is applied to extract clip-level features $\{\mathrm{\bf v}_i\}_{i=1}^{T_c}$ in an untrimmed video. Then, we sample $T_v$ clips and obtain a fixed-length video feature $\mathrm{\bf V} \in \mathbb{R}^{d_v\times {T_v}}$, where $d_v$ represents video feature dimension. Meanwhile, we take advantage of Glove~\cite{pennington2014glove} to map each word $\mathrm{\bf x}^l_i$ in a language query to a vector $\mathrm{\bf q}_i \in \mathbb{R}^{d_l}$ and obtain a query feature $\mathrm{\bf Q} \in \mathbb{R}^{d_l\times {T_l}}$.
\paragraph{Multi-modal interaction}
First of all, temporal convolution layers and GRU layers are applied to map $\mathrm{\bf V}$ and $\mathrm{\bf Q}$ into the same latent space to get corresponding video embedding $\mathrm{\bf \widetilde{V}} \in \mathbb{R}^{d_m\times {T_v}}$ and query embedding $\mathrm{\bf \widetilde{Q}} \in \mathbb{R}^{d_m\times {T_l}}$.
Similar to \cite{zhou2022thinking}, we then apply dual transformer-based encoders, \ie video-centric and query-centric encoders, which include multi-head attention, addition, normalization and Feed Forward Network (FFN). A cross-attention mechanism is adopted to allow for the interaction of multi-modal information. 
In this way, more representative features $\mathrm{\bf \widetilde{V}}^{(L)} \in \mathbb{R}^{d_m\times {T_v}}$, $\mathrm{\bf \widetilde{Q}}^{(L)} \in \mathbb{R}^{d_m\times {T_l}}$ are produced through $L$ multi-modal encoder layers. 
\subsection{Positive Moment Estimation}
\label{PME}

Under single positive labels, the positive moment estimation (PME) module aims to predict multiple unobserved labels as pseudo positives for optimization in the subsequent module. The PME is constructed by two lightweight models, \ie proposal-matching and semantic-reconstruction models.
\paragraph{Proposal matching}
We first apply a proposal-clip operation to divide video feature $\mathrm{\bf \widetilde{V}}^{(L)}$ into multi-scale proposals $\mathrm{\bf F}^v \in \mathbb{R}^{d_m \times C}$, where $C$ is the number of generated proposals. These proposals are then fed into one FFN layer to predict matching scores $\mathrm{\bf s}_{match} \in \mathbb{R}^{C}$ as follows,
\begin{equation}
\mathrm{\bf s}_{match} = \mathrm{sigmoid}(\mathrm{\bf FFN}(\mathrm{\bf F}^{v})).
\label{equ:7}
\end{equation}
During training, supervision labels $\mathrm{\bf \hat z} \in \{1,\emptyset\}^C$ are generated for multi-scale proposals, where only the closest proposal with the observed positive moment is set as 1. In contrast to assuming that unobserved labels are negative, we apply a loss to only observed single positive labels to avoid the false-negative noise. However, models will predict positive for all proposals in this way. Thus, we introduce an Expected Positive Regularization~\cite{cole2021multi} to avoid the trivial solution. The loss function is given by
\begin{equation}
\mathcal{L}_{match}\!=\!-\!\!\sum_{i=1}^C\mathds{1}[\mathrm{\bf\hat z}_i\!=\!1] \log{\mathrm{\bf s}_{match,i}}\!+\!\gamma_1(\sum_{i=1}^C{\mathrm{\bf s}_{match,i}}\!-\!k)^2,
\label{equ:8}
\end{equation}
where $k$ is the expected number of positive moments per video-query pair. This Regularization encourages the number of predicted positives per sample to be close to a constant at the batch level. Researchers~\cite{zhang2021understanding} have observed that neural networks will fit correct labels (informative labels) much faster than wrong labels (uninformative labels). Thus, from a perspective of algorithm optimization, networks will predict unobserved positive moments more readily than negative ones when the number of positives is constrained~\cite{cole2021multi}.

\paragraph{Semantic reconstruction}
To better uncover potential positive moments, we build a lightweight semantic-reconstruction model that reconstructs semantic information of queries from proposal features. Based on a single positive label, a sequential video embedding within its interval is first extracted from the video embedding $\mathrm{\bf \widetilde{V}}$. Then, a random weighted averaging operation is applied to augment $N_s$ fixed-dimensional features $\{\mathrm{\bf f}^v_i\}^{N_s}_{i=1}$. We use an attention-based Recurrent Neural Network~\cite{bahdanau2014neural} as the language generator $\Phi(\cdot)$, which utilizes these features to reconstruct \textit{nouns} and \textit{verbs} of one query. In the optimization phase, a standard captioning loss is used to maximize the normalized log-likelihood of the correct words,
\begin{equation}
\mathcal{L}_{semantic}= -\frac{1}{N_sT_s}\sum_{i=1}^{N_s}\sum_{t=1}^{T_s}\log{p(\mathrm{\bf x}^{l'}_t|\Phi(\mathrm{\bf f}^v_i))},
\label{equ:9}
\end{equation}
where $\mathrm{\bf x}^{l'} $ denotes a \textit{noun}/\textit{verb} of one query and $T_s$ is total time step of the recurrent model. Intuitively, positive moments can better reconstruct the semantic information of a query. In the estimation phase, we leverage this recurrent model to generate words for both multi-scale proposals and the labeled moment, and then compare semantic similarities between them. In our work, we adopt the BLEU-1~\cite{papineni2002bleu} to compute the semantic score $\mathrm{\bf s}_{semantic} \in \mathbb{R}^{C}$.

\paragraph{Positive moment estimation}
With Equation~\ref{equ:8} and Equation~\ref{equ:9}, the PME module is optimized by
\begin{equation}
\mathcal{ L}_{PME}= \mathcal{ L}_{match} + \gamma_2\mathcal{ L}_{semantic}. 
\label{equ:9.5}
\end{equation}

After each epoch, proposals with both $\mathrm{\bf s}_{match}$ and $\mathrm{\bf s}_{semantic}$ below a threshold $T_{thresh}$ are discarded from these multi-scale proposals, which are more likely to be negatives. We further apply Non-Maximal Suppression to filter out highly overlapped proposals and obtain multiple pseudo-positive labels $\{\mathrm{\bf \Hat y}_{se}\}^{N-1}$.
\subsection{Diverse Moment Regression}
\label{DMR}
Similar with the multi-modal encoders, a transformers-based diverse moment regression (DMR) module is built as the decoder. It transforms the video feature $\mathrm{\bf \widetilde{V}}^{(L)}$ along with $N$ learnable input embeddings into $N$ output embeddings $\mathrm{\bf F}_{out}^{(L)} \in \mathbb{R}^{N \times d_m}$, and then produces $N$ predictions $\{\mathrm{\bf b}_{se,i}\}_{i=1}^N$ by FFN layers. To correlate outputs well with semantic information, these input embeddings are encoded by query features $\mathrm{\bf \widetilde{Q}}^{(L)}$. Same with DeNet~\cite{zhou2021embracing}, we add an auxiliary head to predict the center-width $\{\mathrm{\bf b}_{cw,i}\}_{i=1}^N$ to assist temporal grounding. The final predictions are obtained as follows, 
\begin{equation}
\mathrm{\bf b}_{o,i} = \mathrm{\bf FFN}_{o}(\mathrm{\bf F}_{out,i}^{(L)}) \in \mathbb{R}^{2}, \quad o \in \{se,cw\}.
\label{equ:11}
\end{equation}

\begin{table*}[!htp]
\caption{Comparisons in terms of single-label metrics. Bold font indicates best results, underlined second-best.}
\label{tab:1}
\setlength{\tabcolsep}{9pt}
\centering
{
\begin{tabular}{lcccccccc}
\toprule[1pt]
\multirow{3.5}{*}{\bf Method }& \multicolumn{4}{c}{ \bf Charades-STA}& \multicolumn{4}{c}{\bf ActivityNet Captions} \\
\cmidrule(r){2-5}\cmidrule(r){6-9}
 & R@1 & R@1 & R@5 & R@5& R@1 & R@1 & R@5 & R@5 \\
                             & IoU=0.5 & IoU=0.7 & IoU=0.5 & IoU=0.7& IoU=0.3 & IoU=0.5 & IoU=0.3 & IoU=0.5  \\
\midrule
EXCL~\cite{ghosh2019excl} & 44.10 & 22.40 & - & - & 62.30 & 42.70 & -&-\\
TGN~\cite{chen2018temporally} &-&-&-&-& 45.51 & 28.47 & 57.32 &43.33\\
MLVI~\cite{xu2019multilevel} & 35.60 & 15.80 & 79.40 & 45.40&-&-&-&-\\
DEBUG~\cite{lu2019debug} & 37.39 & 17.69 & - &-& 55.91 & 39.72 &-&-\\
VSLNet~\cite{zhang2020span} & 54.19 & 35.22 & - &-& 63.16 & 43.22  &-&-\\
CBP~\cite{wang2020temporally}& 36.80 & 18.87 & 70.94 & 50.19&-&-&-&-\\
2D-TAN~\cite{zhang2020learning} & 42.80 & 23.25 & 80.54 &54.14& 59.45 & 44.51 &85.53 & 77.13\\
DRN~\cite{zeng2020dense} & 53.09 & 31.75 & 89.06 & 60.05 & - & \underline{45.45} & -& \underline{77.97}\\
SCDM~\cite{yuan2020semantic}& 54.44 & 33.43 & 74.43 & 58.08& 54.80 & 36.75 & 77.29 & 64.99\\
CMHN~\cite{hu2021video} &-&-&-&-& 62.49 & 43.47 & 85.37&73.42\\
IVG~\cite{Nan_2021_CVPR} &50.24 &32.88 &-& -& 63.22 &43.84 &-&-\\
MABAN~\cite{sun2021maban} &56.29 &32.26& - & -& - & 44.88 &-& -\\
FVMR~\cite{gao2021fast} &55.01 & 33.74 & \underline{89.17} & 57.24 & 60.63 & 45.00 & 86.11 &77.42 \\
CPN~\cite{Zhao_2021_CVPR} &\underline{59.77} & 36.67& - & -&62.81 &  45.10 &-&-\\
DeNet~\cite{zhou2021embracing} &59.70 & \underline{38.52}& \bf 91.24 & \underline{66.83}&\underline{63.68} & 44.48 & \bf 88.10 &76.91\\
\bf DTG-SPL &\bf 60.05 & \bf 40.13 &  87.34 & \bf 67.12 &\bf 65.18&\bf 47.04&\underline{87.08}&\bf 79.16\\
\bottomrule[1pt]
\end{tabular}
}
\end{table*}
\paragraph{Optimization}
With the Hungarian algorithm, a multi-moment matching operation is performed to search for an optimal permutation index $\bf \hat c$ of $N-1$ elements between pseudo-positive labels $\{\mathrm{\bf \Hat y}_{se}\}^{N-1}$ and predictions $\{(\mathrm{\bf b}_{se,i},\mathrm{\bf b}_{cw,i})\}^N_{i=2}$.
The loss function of DMR is given by
\begin{equation}
\begin{aligned}
\mathcal{L}_{DMR}  & = \mathcal{ L}_{reg}(\mathrm{\bf b}_{se,1};\mathrm{\bf z}_{se}) +  \mathcal{ L}_{reg}(\mathrm{\bf b}_{cw,1};\mathrm{\bf z}_{cw}) \\
+ \mathcal{ L}_{att} & (\mathrm{\bf a}_1;\mathrm{\bf m})+ \lambda \frac{1}{N\!-\!1}  \sum_{i=2}^{N} [\mathcal{ L}_{reg}(\mathrm{\bf b}_{se,i};\mathrm{\bf \hat y}_{se,{\bf \hat c}_{(i)}}) \\
& + \mathcal{ L}_{reg}(\mathrm{\bf b}_{cw,i};\mathrm{\bf \hat y}_{cw,{\bf \hat c}_{(i)}}) +  \mathcal{ L}_{att}(\mathrm{\bf \hat a}_i;\mathrm{\bf \hat m}_{{\bf \hat c}_{(i)}})].
\end{aligned}
\label{equ:13}
\end{equation}

In the Equation~\ref{equ:13}, $\mathrm{\bf \Hat y}_{cw, {\bf \hat c}_{(i)}}$ is converted from $\mathrm{\bf \Hat y}_{se, {\bf \hat c}_{(i)}}$. $\mathcal{ L}_{reg}$ is a regression loss to measure $l_1$ distance, and $\mathcal{ L}_{att}$ is an attention loss~\cite{yuan2019find} to enhance features within groundtruth interval,
\begin{equation}
\mathcal{ L}_{att}(\mathrm{\bf a}_i,\mathrm{\bf m})= -\frac{\sum_{j=1}^{T_v} m_j \mathrm{log} a_{i,j}}{\sum_{j=1}^{T_v} m_j},
\label{equ:14}
\end{equation}
where $\mathrm{\bf a}_i \in \mathbb{R}^{T_v}$ is the attention coefficient of video features $\mathrm{\bf \widetilde{V}}^{(L)}$ from the last self-attention layer in the diverse moment regression module, and $\mathrm{\bf m}$ is the interval mask related to groundtruth. If the j-th clip is within the groundtruth interval, then $m_j = 1 $, otherwise $m_j = 0 $.

\section{Experiments}
\subsection{Datasets and Evaluation Metrics}
\label{metrics}
\paragraph{Datasets}
Experiments are conducted on two popular benchmarks, \ie Charades-STA and ActivityNet Captions. For Charades-STA~\cite{sigurdsson2016hollywood,gao2017tall}, there are 12,408 video-query pairs in the training set and 3,720 pairs in the testing set. For ActivityNet Captions~\cite{caba2015activitynet,krishna2017dense}, 37,417, 17,505, and 17,031 video-query pairs belong to training set, val\_1 set, and val\_2 set. We follow common practices to take val\_2 set as the testing set. Both datasets initially only provide single positive labels. Otani \etal~\cite{otani2020uncovering} recently re-annotate 5 temporal annotations for each pair in the partial testing set, including 1,000 samples on Charades-STA and 1,288 on ActivityNet Captions.

\paragraph{Evaluation metrics}
Two types of metrics, \ie single-label and multi-label metrics, are adopted in our work. The single-label metric, "{\bf R@\textit{N}, IoU=$\alpha$}", is defined as the percentage of single positive labels having at least one close prediction (with IoU larger than $\alpha$) in top-\textit{N} predictions. Typically, $N \in \{1, 5\}$ and $\alpha \in \{0.3, 0.5, 0.7\}$ for different benchmarks. Besides, two multi-label metrics~\cite{zhou2021embracing} are used to meet for the multiple annotations situation. The first one, "{\bf R@(\textit{N, G}), IoU=$\alpha$}", is defined as the percentage of multi-label annotations that match at least one prediction (with IoU larger than $\alpha$) in top-\textit{N} predictions and \textit{G} annotations. The second one, "{\bf R$_\beta$@(\textit{N, G}), IoU=$\alpha$}", ignores low-quality annotations, \ie those in which average IoU among annotations smaller than $\beta$. 
\subsection{Implementation Details}
\label{details}
Following previous works, we adopt I3D features ~\cite{carreira2018action} for video feature extraction, and sample $T_v$=128 video clips. The dimensions $d_v, d_l, d_m$ are set as 1024, 300, 512, respectively. Both encoder and decoder contain $L$= 6 layers. Video features $N_s$ is 5 in Equation~\ref{equ:9} and the domain knowledge $k$ in Equation~\ref{equ:8} is set as 5, equal to the number of predictions $N$. The trade-off parameters $\gamma_1,\gamma_2,\lambda$ are set as 0.1, 0.05, 0.5, respectively.
Same with~\cite{zhang2020learning}, we generate $C$=136 multi-scale proposals for Charades-STA and $C$=1,104 proposals for ActivityNet Captions. $T_{thresh}$ is set as 0.5 to filter out negative proposals. Besides, PME is performed to estimate labels every epoch, and DMR is optimized with only single positive labels in the first epoch. All experiments can be carried out on a single NVIDIA TITAN XP.
\begin{table*}[t]
\caption{Ablation studies of positive moment estimation.}
\label{tab:2}
\setlength{\tabcolsep}{6pt}
\centering
{
\begin{tabular}{lcccccccc}
\toprule[1pt]
\multirow{3.5}{*}{\bf Method }& \multicolumn{4}{c}{ \bf Charades-STA}& \multicolumn{4}{c}{\bf ActivityNet Captions} \\
\cmidrule(r){2-5}\cmidrule(r){6-9}
 & R@5 & R@5 & R@(5,5) & R$_{0.5}$@(5,5)& R@5 & R@5 &  R@(5,5)  & R$_{0.4}$@(5,5) \\
                             & IoU=0.5 & IoU=0.7 & IoU=0.5 & IoU=0.5& IoU=0.3 & IoU=0.5 & IoU=0.5 & IoU=0.5  \\
\midrule
\bf DTG-SPL(full) &\bf87.34 & \bf  67.12 &  \bf 61.88 & \bf 69.79 & \bf 87.08&\bf 79.16 & \bf59.32& \bf 71.11\\
\midrule
w/o reconstruction & 83.63 & 63.28 & 59.10 & 67.27 & 73.45 & 64.40 & 48.13&60.33\\
w/o augmenting & 85.12&64.74&\underline{61.22}&\underline{69.56}&\underline{85.96}&\underline{77.58}&\underline{58.90}&\underline{70.49}\\
\midrule
w/o matching & \underline{86.07} & \underline{66.24} & 60.34 & 68.14&84.94&76.90&58.50&69.80\\
w/o epr & 78.89 & 56.83 & 55.68 &63.94& 70.07& 61.05 &42.10&53.15\\
\bottomrule[1pt]
\end{tabular}
}
\end{table*}
\subsection{Comparison to State-of-the-art Methods}
We compare DTG-SPL with recent methods to validate its effectiveness. First, we evaluate our method using standard single-label metrics in Table\ref{tab:1}. Despite that DTG-SPL does not strive to match only single positive labels, it still achieves superior performance in terms of single-label metrics, and even establishes new state-of-the-art results in most settings. On Charades-STA, DTG-SPL outperforms the recent method DeNet~\cite{zhou2021embracing} by 1.61\% for R@1, IoU=0.7. On ActivityNet Captions, it outperforms DRN~\cite{zeng2020dense} by 1.59\% for R@1, IoU=0.5. This is because single positive labels are used as supervision information to optimize the top-1 output. As a result, DTG-SPL also allows for precise moment localization. Besides, we notice that DTG-SPL does not outperform the previous state-of-the-art in terms of R@5, IoU=0.5 on Charades-STA and R@5, IoU=0.3 on ActivityNet Captions. We consider it might be too loose to match a single label with 5 predictions at low levels of overlap.

\begin{figure}[t]
\begin{center}
   \includegraphics[width=1.0\linewidth]{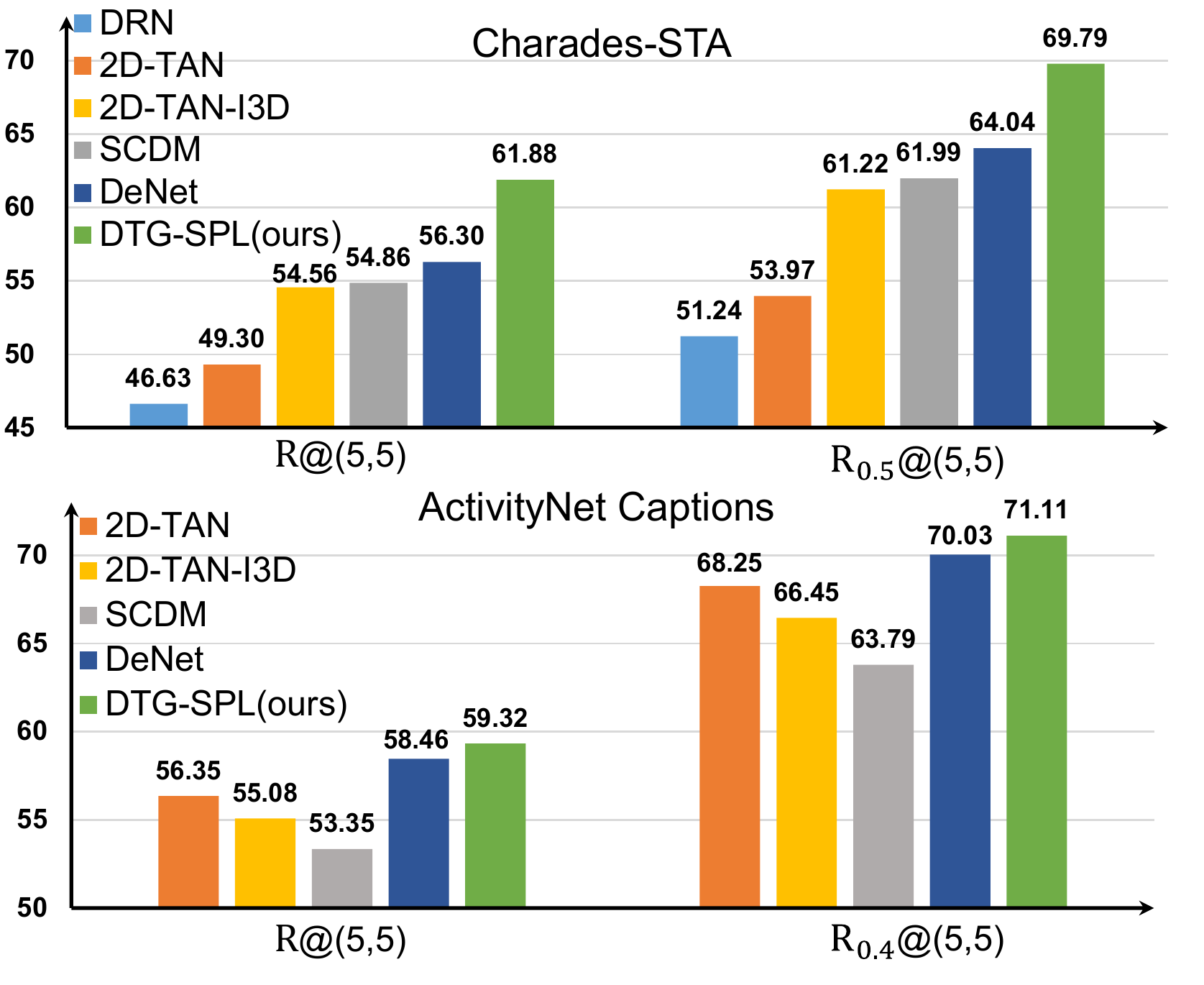}
\end{center}
   \caption{Comparisons in terms of multi-label metrics. The IoU is set as 0.5, and 5 predictions
and 5 annotations are taken into consideration. Best viewed in color.}
\label{fig:3}
\end{figure}

To better reveal the performance of multiple predictions, we further adopt multi-label metrics to evaluate our method. Consistent with single-label metrics and extended annotations, we take at most 5 predictions ($N$ = 5) and 5 temporal annotations ($G$ = 5) in this work. Figure~\ref{fig:3} illustrates the results. DTG-SPL outperforms previous methods (DRN, 2D-TAN, SCDM, DeNet) on both datasets. For example, DTG-SPL yields 5.75\% gains on Charades-STA in terms of "R$_{0.5}$"@(5,5), IoU=0.5", and 1.06\% gains on large-scale ActivityNet Captions in terms of "R$_{0.4}$"@(5,5), IoU=0.5". It means that our proposed DTG-SPL can generate diverse predictions to better meet different users in realistic scenarios.
\begin{figure*}[!th]
\begin{center}
   \includegraphics[width=1.0\linewidth]{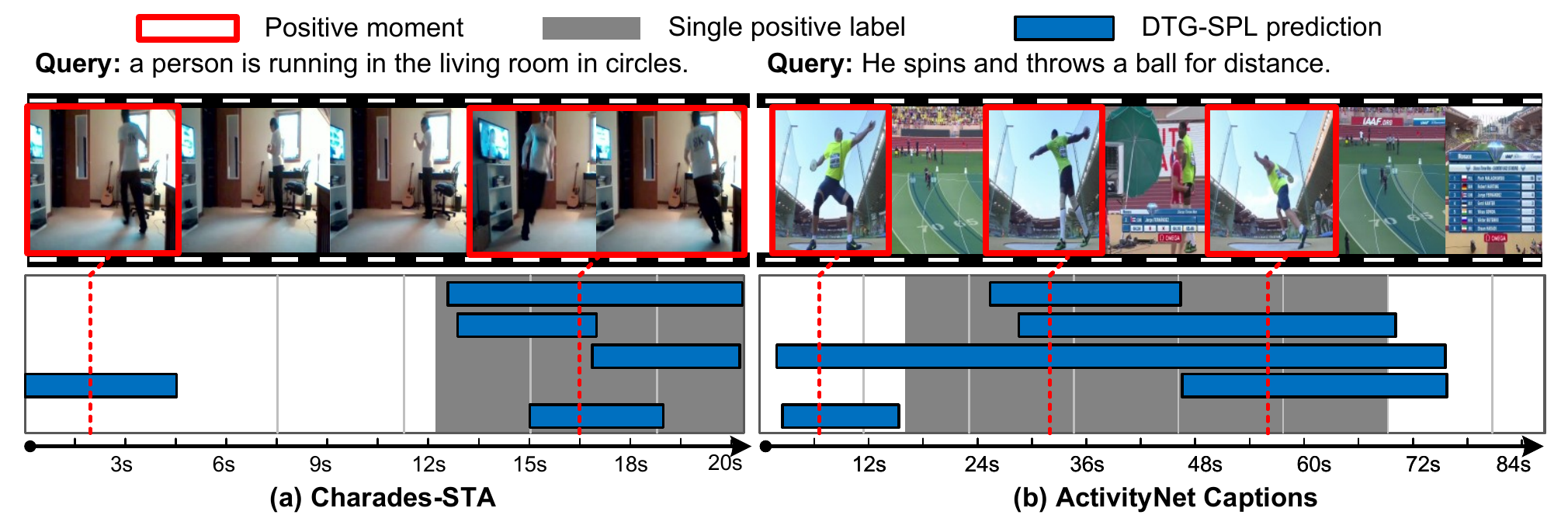}
\end{center}
   \caption{Qualitative results of our proposed DTG-SPL on both Charades-STA and ActivityNet Captions.}
\label{fig:4}
\end{figure*}
\subsection{Ablation Studies}
We first investigate the effects of the semantic-reconstruction model. "DTG-SPL(full)" is our proposed method with full settings, and "w/o reconstruction" indicates the semantic-reconstruction model is removed from the PME module. In our work, we apply a random weighted averaging operation to augment videos features, \ie $N_s$ = 5 video feature available during optimization. "w/o augmenting" discards this augmentation operation, \ie $N_s$ = 1 video feature available. As listed in Table~\ref{tab:2}, we find the semantic-reconstruction model assists in better mining positive moments, and feature augmentation brings further gains.

We then conduct experiments to explore the proposal-matching model. "w/o matching" indicates we remove the proposal-matching model in the PME module, and "w/o epr" abandons the expected positive regularization (dubbed epr) in Equation~\ref{equ:8}. Instead, we turn to the traditional binary cross-entropy loss, where unlabeled moments are considered negatives. This can be regarded as the baseline with the original one-vs-one formulation. From the results in Table~\ref{tab:2}, we can see that 1) positive moment estimation can also benefit from the proposal-matching model, and 2) there is a substantial performance drop without the expected positive regularization, even below "w/o matching". It indicates that assuming unlabeled moments as negatives will introduce false negatives, which in turn hurts the diversity of temporal grounding.

\begin{table}
\caption{Effects of hyper-parameters. $C$ denotes the number of multi-scale proposals, and $k$ is the expected number of positive moments per video-query pair.}
\label{tab:3}
\setlength{\tabcolsep}{5pt}
\centering
{
\begin{tabular}{cccccc}
\toprule[1pt]
\multicolumn{2}{c}{ \multirow{3.5}{*}{Settings}}& \multicolumn{2}{c}{ \bf Charades-STA}& \multicolumn{2}{c}{\bf ActivityNet Captions} \\
\cmidrule(r){3-4}\cmidrule(r){5-6}
 & & R@5 & R@(5,5) & R@5 &  R@(5,5) \\
                             & & IoU=0.7 & IoU=0.5 & IoU=0.5 & IoU=0.5 \\
\midrule
&136&\bf 67.12&61.88&68.62&54.23\\
$C$&428&66.43&\bf 62.64&73.06&57.94 \\
&1,104&67.10&59.96&\bf 79.16& \bf 59.32\\
\midrule
&3&66.46&59.80&74.36&56.78\\
$k$&5&\bf 67.12&\bf 61.88&\bf 79.16& \bf 59.32 \\
&10& 64.65 & 60.34&77.45&58.69 \\
\bottomrule[1pt]
\end{tabular}
}
\end{table}
\subsection{Hyper-Parameters Analysis}
Extended experiments are conducted to analyze the effects of two main hyper-parameters, including the number of multi-scale proposals $C$ and the expected number of positive moments $k$ per video-query pair in Equation~\ref{equ:8}. 

We first divide each video into multi-scale proposals, where the minimum length is \{1/16, 1/32, 1/64\} of the untrimmed video, respectively. A sparse sampling operation~\cite{zhang2020learning} is then employed to eventually yield $C$ = \{136, 428, 1,104\} proposals. The results are listed in Table~\ref{tab:3}. As $C$ increases, we observe performance improvements on ActivityNet Captions, indicating that a larger $C$ could bring more precise proposals. At the same time, we find there are no gains on Charades-STA. This is because more proposals also increase the difficulty of positive moment estimation. Thus, we believe 1/16 resolution is sufficient on Charades-STA due to the relatively short video durations. 

Then, we set $k$ to \{3, 5, 10\} in Table~\ref{tab:3}. We observe that increasing $k$ from 3 to 5 brings improvements, and further increasing leads to drops. It suggests that a too-large $k$ would force networks to incorrectly predict negative instances as positives, which harms final performance. As a result, we set $k$ = 5 in our work, equal to the number of DTG-SPL‘s outputs.

\subsection{Qualitative Performance}
Figure~\ref{fig:4} visualizes multiple predictions generated by our proposed DTG-SPL. The grey area represents the initial single positive label, the red boxes are multiple positive moments provided by~\cite{otani2020uncovering}, and the blue bars are diverse predictions of our DTG-SPL. We can find that the top-5 outputs can cover various positive moments very well. As shown in Figure~\ref{fig:4}.a, the fourth prediction of DTG-SPL localizes the first positive moment in spite of not being labeled. 
It validates the effectiveness of our approach towards diverse temporal grounding under single positive labels.

\section{Conclusion}
\label{conclusion}
In this paper, we reformulate temporal grounding in a condition of single positive labels and put forward a diverse temporal grounding framework, termed DTG-SPL. It leverages an online PME module to uncover positive labels, thereby facilitating a DMR module to localize multiple moments in parallel. Extensive experiments validate DTG-SPL can produce diverse predictions that meet different users in realistic scenarios. 
Since we mine positive moments through multi-scale proposal generation, it would also introduce extra computation costs. In the future, we will explore a more effective and efficient label estimation mechanism.

{\small
\bibliographystyle{ieee_fullname}
\bibliography{main}
}

\end{document}